\documentclass{article}
\usepackage{ijcai26}

\usepackage{times}
\usepackage{soul}
\usepackage{url}
\usepackage[hidelinks]{hyperref}
\usepackage[utf8]{inputenc}
\usepackage[small]{caption}
\usepackage{graphicx}
\usepackage{subcaption}
\usepackage{cuted}
\usepackage{svg}
\usepackage{amsmath}
\usepackage{amsthm}
\usepackage{amssymb}
\usepackage{booktabs}
\usepackage{algorithm}
\usepackage{algorithmic}
\usepackage[switch]{lineno}
\usepackage{xcolor}

\usepackage{multirow}
\usepackage[table]{xcolor}
\newcolumntype{P}[1]{>{\centering\arraybackslash}p{#1}}

\newtheorem{proposition}{Proposition}

\title{Not All Timesteps Matter Equally: 
Selective Alignment Knowledge Distillation for Spiking Neural Networks}

\author{
Kai Sun$^1$
\and
Peibo Duan$^{1*}$\and
Yongsheng Huang$^2$\and
Guowei Zhang$^1$\and
Benjamin Smith$^1$\and
Nanxu Gong$^3$\And
Levin Kuhlmann$^{1}$\thanks{Corresponding author.}\\
\affiliations
$^1$Faculty of Information Technology, Monash University, Australia\\
$^2$School of Software, Northeastern University, China\\
$^3$Department of Medicine, National University of Singapore, Singapore\\
\emails
\{kai.sun1, peibo.duan, levin.kuhlmann\}@monash.edu
}

\begin{document}

\maketitle

\begin{abstract}
Spiking neural networks (SNNs), which are brain-inspired and spike-driven, achieve high energy efficiency. However, a performance gap between SNNs and artificial neural networks (ANNs) still remains. Knowledge distillation (KD) is commonly adopted to improve SNN performance, but existing methods typically enforce uniform alignment across all timesteps, either from a teacher network or through inter-temporal self-distillation, implicitly assuming that per-timestep predictions should be treated equally. In practice, SNN predictions vary and evolve over time, and intermediate timesteps need not all be individually correct even when the final aggregated output is correct. Under such conditions, effective distillation should not force every timestep toward the same supervision target, but instead provide corrective guidance to erroneous timesteps while preserving useful temporal dynamics. To address this issue, we propose \textbf{S}elective \textbf{A}lignment \textbf{K}nowledge \textbf{D}istillation \textbf{(SeAl-KD)}, which selectively aligns class-level and temporal knowledge by equalizing competing logits at erroneous timesteps and reweighting temporal alignment based on confidence and inter-timestep similarity. Extensive experiments on static image and neuromorphic event-based datasets demonstrate consistent improvements over existing distillation methods. The code is available at \url{https://github.com/KaiSUN1/SeAl}.
\end{abstract}

\section{Introduction}
Spiking Neural Networks (SNNs), often regarded as the third generation of neural networks, are biologically inspired models that communicate through discrete spikes rather than continuous-valued activations (\cite{maass1997networks}). By processing information in an event-driven manner with sparse spikes over time, SNNs offer the potential for high energy efficiency, especially when deployed on neuromorphic hardware (\cite{JointA-SNN}). Nevertheless, training SNNs to achieve accuracy comparable to Artificial Neural Networks (ANNs) remains challenging, because useful evidence is accumulated progressively through spikes, making different timesteps unevenly informative during learning (\cite{bellec2018long,neftci2019surrogate}).

To narrow the performance gap between ANNs and SNNs, knowledge distillation (KD) has become a widely adopted strategy (\cite{KDSNN,SAKD}). Early distillation methods typically supervise only the final aggregated output by matching averaged logits, which may mix inconsistent temporal information during optimization (\cite{alltimeconsistent,TET}). More recent approaches move toward timestep-wise distillation by injecting teacher supervision at each timestep and aligning logits throughout the temporal dimension (\cite{ensemble1,ensemble2}). In parallel, several methods further exploit the temporal structure of SNNs through self-distillation (\cite{synergy}) or temporal consistency regularization (\cite{alltimeconsistent,adtimeconsistent}), encouraging predictions or features from different timesteps to be consistent in order to stabilize optimization and improve performance (\cite{synergy,ensemble1}). 

\begin{figure}[t]
    \centering
    \includegraphics[width=1\linewidth]{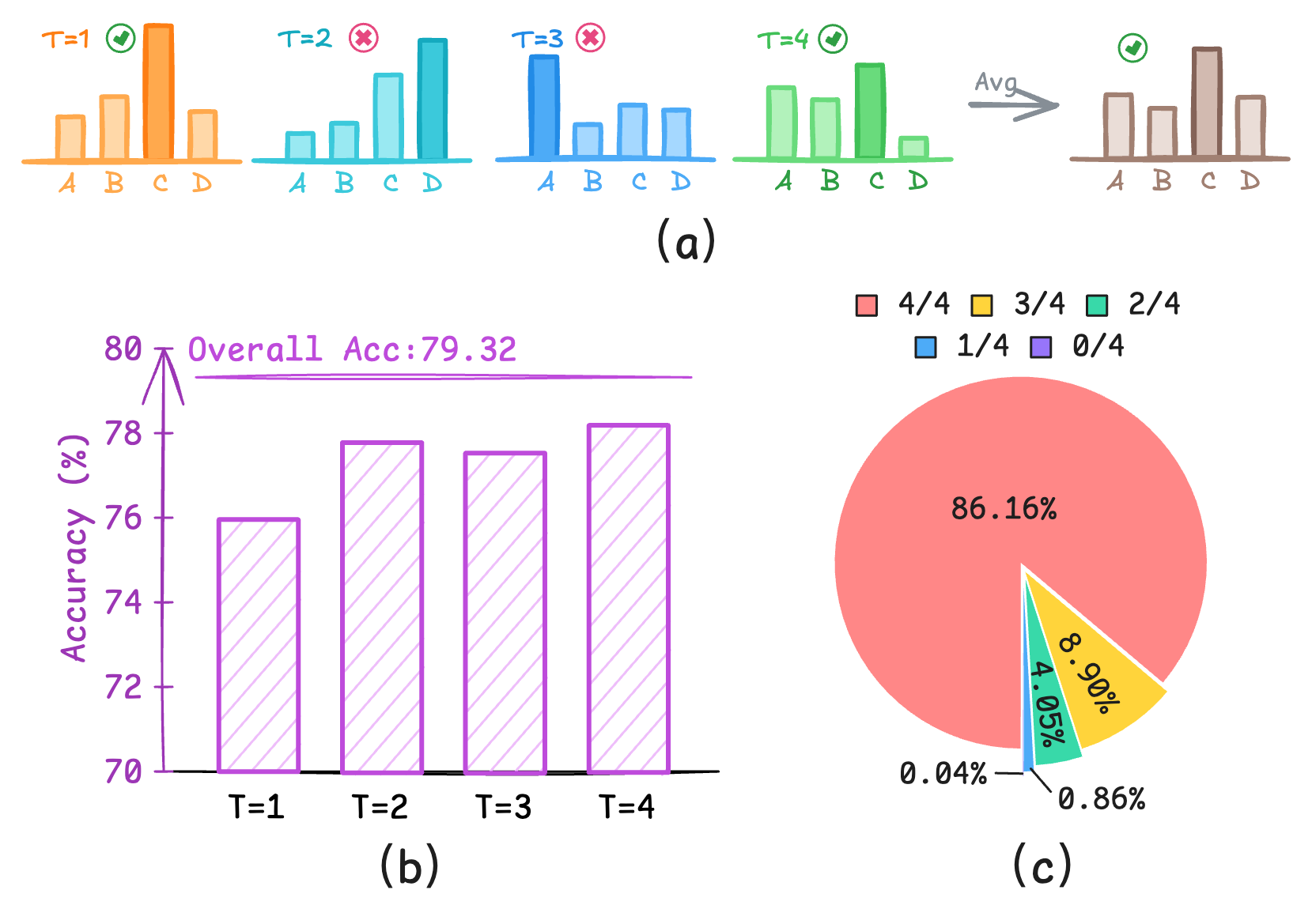}
    \caption{
    Mismatch between intermediate and final predictions in SNNs. (a) A toy example where intermediate predictions are wrong, but the final aggregated prediction is correct. (b) Per-timestep accuracy and final aggregated accuracy on CIFAR100. (c) Distribution of CIFAR100 samples with correct final predictions by the number of correctly predicted timesteps out of $T$.
    }
    \label{fig:intro}
\end{figure}

The consistency assumption is partially misaligned with the intrinsic properties of SNNs and their prediction mechanism. Due to membrane potential integration and reset, spike firing may induce abrupt changes in membrane states, making it difficult for SNNs to maintain identical predictions across all timesteps, as detailed in \textbf{Appendix~\ref{app:lif_variance}}. Meanwhile, since the final decision is determined by temporal accumulation rather than any single timestep, an incorrect intermediate prediction does not necessarily imply an incorrect final outcome. As illustrated in the toy case in Figure~\ref{fig:intro}(a), some intermediate timesteps are misclassified while the final temporally aggregated prediction remains correct. Figure~\ref{fig:intro}(b) further shows that per-timestep accuracy is consistently lower than temporally aggregated accuracy. Moreover, Figure~\ref{fig:intro}(c) reveals that more than 12\% of correctly classified samples are misclassified at some intermediate timesteps, and some samples are never correctly classified at any individual timestep but still become correct after temporal aggregation. These observations suggest that a timestep should not be judged solely by whether it is already correct, but by whether it contributes useful evidence to the final temporal accumulation.

However, existing timestep-wise distillation strategies do not explicitly account for this distinction. By encouraging each timestep to align with the teacher signal, they may overlook what corrective evidence is actually needed for improving temporal accumulation and which temporal sources can provide reliable support. This perspective leads to two key questions for timestep-wise distillation: \textbf{what corrective evidence should be injected into a currently erroneous timestep}, and \textbf{from which temporal sources should it be drawn}? For a currently erroneous timestep dominated by a confusing wrong class, directly transferring teacher preference over the full class distribution may dilute the needed correction and interfere with the adjustment of the most critical class relation, namely that between the ground-truth class and the wrongly favored class. Moreover, in temporal self-distillation, not all source timesteps are equally reliable: some provide confident and compatible evidence, while others may introduce noisy or misleading temporal signals. To address these issues, we propose \textbf{SeAl-KD}, a selective distillation framework with two components. \textbf{Error-aware Logit Alignment (ELA)} refines the class evidence received by erroneous timesteps, while \textbf{Selective Temporal Alignment (STA)} emphasizes reliable and compatible source timesteps during alignment.

\begin{figure*}[t]
    \centering
    \includegraphics[width=\textwidth]{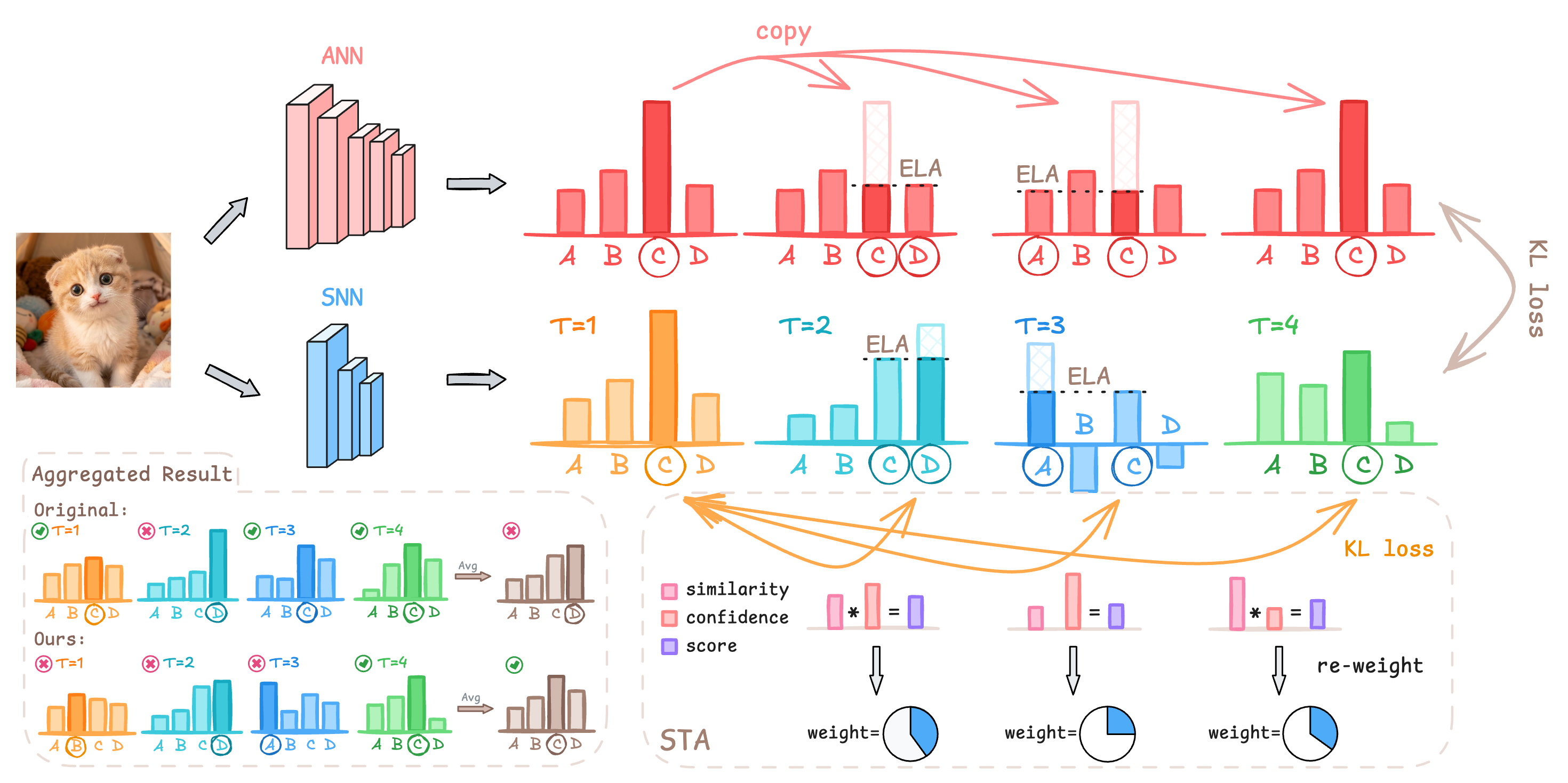}
    \caption{SeAl-KD framework. SNNs learn from the same copied ANN output across timesteps. ELA equalizes the true and predicted-false logits at erroneous timesteps before teacher–student KL. STA reweights temporal KL using confidence and inter-timestep similarity. $C$ denotes the ground-truth class. Left-bottom illustration shows that our method aims to follow temporal diversity to obtain correct predictions, even when some intermediate predictions are incorrect.}
    \label{fig:framework}
\end{figure*}

Our main contributions are summarized as follows:
(1) We reveal that erroneous timesteps are prevalent in SNNs and show that intermediate misclassifications do not necessarily impair the final prediction under temporal aggregation, exposing the mismatch between consistency and temporal evidence accumulation.
(2) We propose SeAl-KD, a selective KD framework that improves corrective supervision for erroneous timesteps through ELA and STA. We further provide theoretical analysis to justify the proposed selective alignment.
(3) We conduct extensive experiments on both static and neuromorphic image datasets. The results show that SeAl-KD preserves richer temporal distributions across timesteps and consistently improves performance.

\section{Related Works}
\subsection{Knowledge Distillation for SNNs}
KD improves SNN training by transferring final outputs, logits, or features from ANNs to narrow the gap with real-valued ANNs (\cite{KDSNN,LaSNN}). Along this line, distillation strategies have evolved from global supervision on temporally aggregated outputs (\cite{HTA-KL}) to timestep-wise supervision that aligns logits or representations at each timestep (\cite{ensemble1,ensemble2}). In addition, self-distillation and temporal consistency regularization further exploit the student’s temporal structure to align outputs across timesteps (\cite{SAKD,TSSD}). Despite these advances, most existing methods impose supervision on different timesteps in a largely uniform manner, without distinguishing whether a timestep is currently erroneous, what corrective information it actually needs, or whether the temporal source used for supervision is reliable.

\subsection{Temporal Discrepancy in SNNs}
Temporal discrepancy is intrinsic to SNNs, as membrane integration and spike-triggered resets cause neuronal states and predictions to evolve across timesteps (\cite{bellec2018long}). Prior work has improved temporal modeling and training stability through learnable membrane dynamics (\cite{plif}), temporal normalization (\cite{STBP-tdBN,duan2022temporal}), and surrogate-gradient design (\cite{li2021differentiable,wang2023adaptive}). Some studies further exploit temporal structure via timestep-dependent weighting (\cite{TET}) or temporal consistency regularization across timesteps (\cite{alltimeconsistent,adtimeconsistent}). However, these methods mostly regulate temporal discrepancy uniformly, without explicitly considering the role of each timestep in the final temporally accumulated prediction. Consequently, they do not distinguish what supervision a timestep needs or which temporal sources are reliable for providing it.

\section{Preliminaries}
\paragraph{Timestep-wise distillation.} Consider an SNN unrolled over $T$ timesteps. At timestep $t$, the student SNN model, denoted by the superscript $S$, produces a logit vector $\mathbf{z}_t^{S} \in \mathbb{R}^C$, where $C$ denotes the number of classes. The final prediction is typically obtained by temporally averaging the logits over all timesteps, while each $\mathbf{z}_t^{S}$ can also be treated as an intermediate prediction. For the same input, the teacher ANN model, denoted by the superscript $A$, produces a time-invariant logit vector $\mathbf{z}^{A}$. The corresponding categorical distributions are given by the temperature-scaled softmax functions $\mathbf{p}_t^{S}=\mathrm{softmax}(\mathbf{z}_t^{S}/\tau)$ and $\mathbf{p}^{A}=\mathrm{softmax}(\mathbf{z}^{A}/\tau)$, where $\tau>0$ is the temperature. The probability of class $i$ under $\mathbf{p}_t^{S}$ and $\mathbf{p}^{A}$ is denoted by $p_{t,i}^{S}$ and $p_i^{A}$, respectively.

For each timestep $t$, the student is optimized with two objectives. The first objective $\mathcal{L}_{\mathrm{CLS}}$ applies cross-entropy (CE) loss for classification supervision at each timestep:
\begin{equation}
\label{eq:cls}
\mathcal{L}_{\mathrm{CLS}}
= \frac{1}{T}\sum_{t=1}^{T}\ell_{\mathrm{CE}}(\mathbf{z}_t^{S}, \mathbf{y})
= \frac{1}{T}\sum_{t=1}^{T}\Big(-\sum_{i=1}^{C} \mathbf{y}_i \log p_{t,i}^{S}\Big),
\end{equation}
where $\mathbf{y}\in\{0,1\}^C$ is the one-hot ground-truth label.

The second objective $\mathcal{L}_{\mathrm{KD}}$ aligns each temporal output with the same teacher distribution using the Kullback--Leibler (KL) divergence:
\begin{equation}
\mathcal{L}_{\mathrm{KD}}
= \frac{1}{T}\sum_{t=1}^{T}\mathrm{KL}\!\big(\mathbf{p}^{A}\,\|\,\mathbf{p}_t^{S}\big)
= \frac{1}{T}\sum_{t=1}^{T}\sum_{i=1}^{C} p_{i}^{A}\log\frac{p_{i}^{A}}{p_{t,i}^{S}}.
\end{equation}

The overall training objective $\mathcal{L}$ is defined as
\begin{equation}
\mathcal{L} = \mathcal{L}_{\mathrm{CLS}} + \lambda \mathcal{L}_{\mathrm{KD}},
\end{equation}
where $\lambda$ controls the contribution of the distillation term.

\section{Methodology}
This section presents SeAl-KD, which consists of ELA and STA. As illustrated in Figure~\ref{fig:framework}, ELA refines class-level correction at erroneous timesteps, while STA selects confident and compatible temporal sources for alignment.

\subsection{Error-aware Logits Alignment Distillation}
ELA performs timestep-wise distillation while accounting for prediction errors across timesteps in SNNs. Let $y^\ast\in\{1,\dots,C\}$ denote the ground-truth class index corresponding to the one-hot label $\mathbf{y}$, and let $z_{t,c}^{S}$ and $z_{t,c}^{A}$ denote the student and teacher logits for class $c$ at timestep $t$, respectively. The student prediction at timestep $t$ is $c_t^{\mathrm{pred}} = \arg\max_{c} z_{t,c}^{S}$. When an intermediate timestep is erroneous, ELA relaxes distillation on its dominant confusion instead of directly enforcing the correct class ordering, avoiding misleading correction on the confusing class pair while preserving supervision on the remaining classes.

\paragraph{Logit modification.}
If $c_t^{\mathrm{pred}} \neq y^\ast$, we focus on the class pair formed by the ground-truth class $y^\ast$ and the predicted false class $c_t^{\mathrm{false}} = c_t^{\mathrm{pred}}$. We equalize only the logits of this pair and keep the remaining classes unchanged. Specifically, the logits of $y^\ast$ and $c_t^{\mathrm{false}}$ are both set to the minimum of their original values, which avoids introducing unnecessary absolute shifts. The modified student and teacher logits, denoted by $\tilde z_{t,c}^{S}$ and $\tilde z_{t,c}^{A}$, are defined as
\begin{equation}
\tilde z_{t,c}^{S} =
\begin{cases}
\min\!\left(z_{t,y^\ast}^{S},\, z_{t,c_t^{\mathrm{false}}}^{S}\right),
& c \in \{y^\ast,\, c_t^{\mathrm{false}}\},\\
z_{t,c}^{S}, & \text{otherwise}.
\end{cases}
\end{equation}
\begin{equation}
\tilde z_{t,c}^{A} =
\begin{cases}
\min\!\left(z_{t,y^\ast}^{A},\, z_{t,c_t^{\mathrm{false}}}^{A}\right),
& c \in \{y^\ast,\, c_t^{\mathrm{false}}\},\\
z_{t,c}^{A}, & \text{otherwise}.
\end{cases}
\end{equation}

\paragraph{ELA objective.}
The ELA loss is then defined as
\begin{equation}
\label{eq:ELA}
\mathcal{L}_{\mathrm{ELA}}
= \frac{1}{T}\sum_{t=1}^{T}
\mathrm{KL}\!\big(p(\tilde{\mathbf{z}}_t^{A})\,\|\,p(\tilde{\mathbf{z}}_t^{S})\big).
\end{equation}
By removing the forced preference between the ground-truth class and the dominant false class at erroneous timesteps, ELA prevents distillation from reinforcing misleading intermediate ordering while retaining teacher guidance on the remaining classes.

\subsection{Similarity-aware Temporal Alignment Distillation}
Based on uniform temporal alignment (\textbf{UTA}) \cite{alltimeconsistent}, which enforces unweighted pairwise alignment across timesteps, STA adopts a weighted temporal distillation scheme that allows each timestep $t$ to selectively learn from confident and compatiable timesteps. This guides weak timesteps toward better temporal states that support effective temporal evidence accumulation.

\paragraph{Timestep confidence.}
The reliability of timestep $t$ is quantified by an entropy-based confidence score. The entropy $H_t$ measures the uncertainty of the class distribution and is normalized by the maximum entropy $\log C$:
\begin{equation}
\mathrm{Conf}_t = 1-\frac{H_t}{\log C},\qquad
H_t = -\!\sum_{c=1}^{C} p(\mathbf{z}_t^{S})_c \log p(\mathbf{z}_t^{S})_c.
\end{equation}
A lower entropy leads to a larger $\mathrm{Conf}_t$, indicating that the timestep provides a more reliable source.

\paragraph{Timestep compatibility.}
The compatibility between a target timestep $t$ and a source timestep $t'$ is measured by cosine similarity in the student logit space. This similarity reflects the consistency of their class-level preferences:
\begin{equation}
\mathrm{Sim}(t, t') =
\frac{\mathbf{z}_t^{S} \cdot \mathbf{z}_{t'}^{S}}
{\|\mathbf{z}_t^{S}\| \,\|\mathbf{z}_{t'}^{S}\|}.
\end{equation}
A larger $\mathrm{Sim}(t,t')$ indicates that the two timesteps exhibit more compatible class preferences.

\paragraph{Source weighting.}
For each target timestep $t$, an unnormalized source score $s_{t,t'}$ is computed for source timestep $t'$ by combining its confidence with its compatibility to the target:
\begin{equation}
s_{t,t'} = \mathrm{Conf}_{t'} \cdot \mathrm{Sim}(t,t').
\end{equation}
The final weights are obtained by applying a softmax over all source timesteps satisfying $t'\neq t$, with $w_{t,t}=0$:
\begin{equation}
w_{t,t'} =
\frac{\exp(s_{t,t'})}{\sum_{j \neq t} \exp(s_{t,j})},
\quad t'\neq t.
\end{equation}
Thus, each target timestep primarily learns from confident and compatible source timesteps.

\paragraph{STA objective.}
Based on the obtained weights, we encourage each target timestep to align with other source timesteps through a weighted KL divergence. The STA loss is defined as
\begin{equation}
\label{eq:STA}
\mathcal{L}_{\mathrm{STA}}
=
\frac{1}{T}\sum_{t=1}^{T}
\sum_{t' \neq t} w_{t,t'}\,
\mathrm{KL}\!\big(p(\mathbf{z}_{t'}^{S})\,\|\,p(\mathbf{z}_{t}^{S})\big).
\end{equation}
This objective lets each timestep absorb selectively weighted temporal guidance, instead of being uniformly aligned to all other timesteps, thereby providing more effective support for weak timesteps and reducing interference on already reliable ones.

\begin{figure*}[t]
    \centering
    \includegraphics[width=0.85\textwidth]{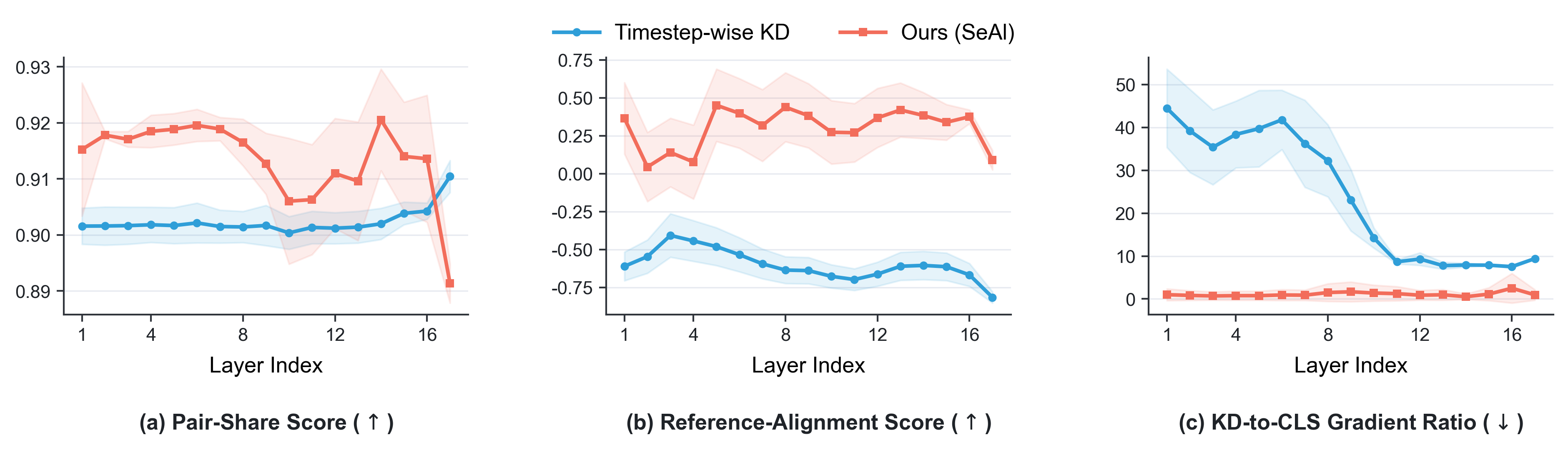}
    \caption{Layer-wise statistics over all timesteps for the three propositions: (a) the fraction of the ELA update assigned to the ground-truth class and the dominant false class at erroneous timesteps; (b) the cosine similarity between the STA update and the direction that reduces the gap to reliability-weighted temporal references at weak timesteps; (c) the ratio between the distillation-gradient norm and the classification-gradient norm at already-correct timesteps. Statistics are computed from five randomly selected samples and reported as mean $\pm$ std.}
    \label{fig:analysis}
\end{figure*}

\begin{table*}[t]
\centering
\begin{tabular}{c c c c c c c}
\toprule
 & \multirow{2}{*}{\raisebox{-0.5ex}{\textbf{Method}}}
 & \multirow{2}{*}{\raisebox{-0.5ex}{\textbf{Architecture}}}
 & \multicolumn{2}{c}{\textbf{CIFAR-10}}
 & \multicolumn{2}{c}{\textbf{CIFAR-100}} \\
\cmidrule(lr){4-7}
 &  &  & T=4 & T=6 & T=4 & T=6 \\
\midrule

\multirow{8}{*}{\parbox[c]{0.5cm}{\centering\rotatebox[origin=c]{90}{w/o KD}}}
& Dspike \cite{Dspike}     & \multirow{3}{*}{ResNet-18} & 93.66 & 94.25 & 73.35 & 74.24 \\
& GLIF \cite{GLIF}      &                            & 94.67 & 94.88 & 76.42 & 77.28 \\
& RateBP \cite{RateBP}    &                            & 95.61 & 95.90  & 78.26 & 79.02 \\
\cmidrule(lr){2-7}

& STBP-tdBN \cite{STBP-tdBN}  & \multirow{5}{*}{ResNet-19} & 92.92 & 93.16 & --    & --    \\
& TET \cite{TET}       &                            & 94.44 & 94.50  & 74.47 & 74.72 \\
& GLIF \cite{GLIF}      &                            & 94.85 & 95.03 & 77.05 & 77.35 \\
& LSG \cite{LSG} &                            & 95.17 & 95.52 & 76.85 & 77.13 \\
& RateBP  \cite{RateBP}   &                            & 96.26 & 96.36 & 80.71 & 80.83 \\
\midrule

\multirow{9}{*}{\parbox[c]{0.5cm}{\centering\rotatebox[origin=c]{90}{w/ KD}}}
& KDSNN \cite{KDSNN}          & \multirow{5}{*}{ResNet-18} & 93.41 & --    & --    & --    \\
& Joint A-SNN  \cite{JointA-SNN}   &                            & 95.45 & --    & 77.39 & --    \\
& Rate-based KD \cite{rate-based-KD} &                            & 95.92 & 96.14 & 78.85 & 79.40  \\
& Logit-SNN \cite{ensemble2} &                            & 95.57 & 95.96 & 79.10  & 79.80  \\
& \cellcolor{pink!25}\textbf{SeAl-KD (Ours)}
&
& \cellcolor{pink!25}\textbf{95.88$_{\pm 0.12}$}
& \cellcolor{pink!25}\textbf{96.18$_{\pm 0.06}$}
& \cellcolor{pink!25}\textbf{79.88$_{\pm 0.13}$}
& \cellcolor{pink!25}\textbf{80.25$_{\pm 0.12}$} \\
\cmidrule(lr){2-7}

& SAKD  \cite{SAKD}          & \multirow{4}{*}{ResNet-19} & 96.06 & --    & 80.10  & --    \\
& HTA-KL \cite{HTA-KL}         &                            & 96.76 & --    & 81.03 & --    \\
& Logit-SNN \cite{ensemble2}  &                            & 96.97 & 97.00    & 82.47 & 82.56 \\
& \cellcolor{pink!25}\textbf{SeAl-KD (Ours)}
& 
& \cellcolor{pink!25}\textbf{97.14$_{\pm 0.06}$}
& \cellcolor{pink!25}\textbf{97.23$_{\pm 0.05}$}
& \cellcolor{pink!25}\textbf{83.04$_{\pm 0.05}$}
& \cellcolor{pink!25}\textbf{83.37$_{\pm 0.08}$} \\
\bottomrule
\end{tabular}
\caption{Comparison of different direct-training and distillation methods on CIFAR-10 and CIFAR-100.}
\label{tab:cifar}
\end{table*}

\subsection{Selective Alignment Distillation}
SeAl-KD combines ELA in Eq.~(\ref{eq:ELA}) and STA in Eq.~(\ref{eq:STA}) to realize selective temporal supervision. Specifically, ELA regulates what class-level knowledge is injected at erroneous timesteps, while STA guides weak timesteps toward weighted better temporal states, so that temporal correction better supports the final evidence accumulation process. Together with the classification loss, the overall training objective is
\begin{equation}
\mathcal{L}_{\mathrm{SeAl\text{-}KD}}
=
\mathcal{L}_{\mathrm{CLS}}
+
\alpha\,\mathcal{L}_{\mathrm{ELA}}
+
\beta\,\mathcal{L}_{\mathrm{STA}},
\end{equation}
where $\alpha$ and $\beta$ control the contributions of ELA and STA, respectively.

\subsection{Theoretical and Statistical Analysis}
We analyze SeAl-KD from the perspective of the additional update introduced by distillation beyond the standard cross-entropy objective. The analysis focuses on three representative timestep conditions: erroneous timesteps, weak timesteps, and already-correct timesteps. For space efficiency, in the main paper we select one timestep for each condition. The full results over additional sampled timesteps are provided in \textbf{Appendix~\ref{app:theoretical_figures}}. These diagnostics show that SeAl-KD intervenes selectively: it localizes correction when the prediction is wrong, transfers temporal guidance from more reliable temporal states, and remains restrained when the prediction is already reliable.

\begin{proposition}[Localized correction]
At an erroneous timestep, ELA encourages the distillation update to concentrate on the error-relevant class subspace, rather than spreading the correction uniformly over all classes.
\end{proposition}

Consider an erroneous timestep $t$ with ground-truth class index $y^\ast$ and currently predicted wrong class $c_t^{\mathrm{false}}$. Let $\bar z_{t,\mathrm{rest}}^{S}$ denote the mean student logit over the remaining classes outside $\{y^\ast,c_t^{\mathrm{false}}\}$, and let $\nabla_{\theta_l}$ denote the gradient with respect to the parameters of layer $l$. To examine whether the ELA update is concentrated on the dominant confusion pair, we define three directional effects:
\begin{equation}
\begin{aligned}
D_{\mathrm{true}}^l &=
\left|\left\langle 
\nabla_{\theta_l}\mathcal{L}_{\mathrm{ELA}},
\nabla_{\theta_l}z_{t,y^\ast}^{S}
\right\rangle\right|,\\
D_{\mathrm{false}}^l &=
\left|\left\langle 
\nabla_{\theta_l}\mathcal{L}_{\mathrm{ELA}},
\nabla_{\theta_l}z_{t,c_t^{\mathrm{false}}}^{S}
\right\rangle\right|,\\
D_{\mathrm{rest}}^l &=
\left|\left\langle 
\nabla_{\theta_l}\mathcal{L}_{\mathrm{ELA}},
\nabla_{\theta_l}\bar z_{t,\mathrm{rest}}^{S}
\right\rangle\right|.
\end{aligned}
\end{equation}
The pair-concentration score is then defined as
\begin{equation}
\mathrm{PairShare}_t^l =
\frac{D_{\mathrm{true}}^l+D_{\mathrm{false}}^l}
{D_{\mathrm{true}}^l+D_{\mathrm{false}}^l+D_{\mathrm{rest}}^l}.
\end{equation}

The inner products measure how strongly an update along the ELA direction acts on each logit. A larger $\mathrm{PairShare}_t^l$ means that the update is more concentrated on the confusing pair $\{y^\ast,c_t^{\mathrm{false}}\}$ than on the remaining classes. As shown in Figure~\ref{fig:analysis}(a), SeAl-KD gives a higher pair-share score than timestep-wise KD, suggesting that ELA yields more localized correction at erroneous timesteps.

\begin{proposition}[Reference-guided correction]
For a weak timestep, STA encourages the update to reduce its margin discrepancy to reliability-weighted temporal references, thereby providing guidance from confident and compatible states.
\end{proposition}

Let $m_t = z_{t,y^\ast}^{S} - z_{t,c_t^{\mathrm{false}}}^{S}$ denote the student margin at timestep $t$, where the margin measures the logit gap between the ground-truth class $y^\ast$ and the dominant false class $c_t^{\mathrm{false}}$. A larger margin indicates that the student assigns stronger relative preference to the ground-truth class over the confusing false class. Let $m_{\mathrm{ref},t}$ be the reference margin for timestep $t$, obtained by aggregating source-timestep margins according to the STA weights. We measure whether the STA update is aligned with the direction that reduces the discrepancy between $m_t$ and $m_{\mathrm{ref},t}$:
\begin{equation}
\mathrm{RefAlign}_t^l=
\cos\!\Big(
\nabla_{\theta_l}\mathcal{L}_{\mathrm{STA}},
\nabla_{\theta_l}(m_t-m_{\mathrm{ref},t})^2
\Big).
\end{equation}

A larger $\mathrm{RefAlign}_t^l$ indicates stronger alignment with the discrepancy-reduction direction, suggesting that STA guides weak timesteps toward reliability-weighted temporal references. Rather than forcing each intermediate timestep to be independently correct, STA promotes more consistent evidence accumulation across timesteps. As shown in Figure~\ref{fig:analysis}(b), SeAl-KD produces positive alignment, whereas timestep-wise KD shows weaker or negative alignment.

\begin{proposition}[Restrained correction]
When a timestep is already correctly classified, the distillation update remains small relative to the task gradient, reducing unnecessary interference with reliable predictions.
\end{proposition}

For already-correct timesteps, a large distillation update may perturb a reliable prediction. We therefore measure the relative strength of the distillation gradient against the task gradient:
\begin{equation}
\mathrm{KDRatio}_t^l=
\frac{\|\nabla_{\theta_l}\mathcal{L}_{\mathrm{KD}}\|}
{\|\nabla_{\theta_l}\mathcal{L}_{\mathrm{CLS}}\|}.
\end{equation}

A smaller $\mathrm{KDRatio}_t^l$ indicates weaker interference from the distillation objective. As shown in Figure~\ref{fig:analysis}(c), SeAl-KD yields a much smaller KD-to-CLS gradient ratio than timestep-wise KD at the representative correct timestep, indicating that it remains restrained when the prediction is already reliable.


\section{Experiments}
\subsection{Implementation Details}
\paragraph{Datasets.} We evaluate the proposed method on four benchmark datasets, including three static image datasets, namely CIFAR-10, CIFAR-100, and ImageNet, as well as one neuromorphic dataset, DVS-CIFAR10. This selection allows us to assess the effectiveness of the proposed approach on both frame-based vision tasks and event-based neuromorphic data. Dataset details are provided in the \textbf{Appendix~\ref{app:dataset-details}}.

\paragraph{Training Settings.} All experiments are implemented in PyTorch and trained on NVIDIA A100 GPUs. The SNNs are built with leaky integrate-and-fire neurons and trained using surrogate gradient learning \cite{neftci2019surrogate}. All experiments are conducted three times and the average results are reported, except for ImageNet. For neuromorphic datasets, ANN training uses the average value of the event data as input. Training details are provided in the \textbf{Appendix~\ref{app:training-details}}.

\subsection{Performance Comparison}
The results on CIFAR-10, CIFAR-100, ImageNet, and DVS-CIFAR10 are summarized in Tables~\ref{tab:cifar}, \ref{tab:Imagenet}, and \ref{tab:DVS}. Across all evaluated benchmarks, SeAl-KD consistently outperforms directly trained SNNs and achieves superior or competitive performance compared with representative logit-based SNN distillation methods and other distillation baselines under comparable architectures and inference timesteps. The improvements are observed on both static frame-based datasets and event-based neuromorphic datasets, indicating that the proposed method generalizes well across different data modalities. In addition, SeAl-KD only introduces lightweight training-time objectives, and a detailed energy analysis is provided in \textbf{Appendix~\ref{app:energy}}.

\begin{table}[t]
\centering
\begin{tabular}{c c c}
\toprule
 & \textbf{Method} & \textbf{Acc(\%)} \\
\midrule
\multirow{5}{*}{\rotatebox{90}{w/o KD}} 
 & TET \cite{TET}       & 68.00 \\
 & Dspike \cite{Dspike}    & 68.19 \\
 & GLIF   \cite{GLIF}    & 67.52 \\
 & RecDis-SNN \cite{RecDis-SNN} & 67.33 \\
 & RateBP  \cite{RateBP}   & 70.01 \\
\cmidrule(lr){1-3}
\multirow{5}{*}{\rotatebox{90}{w/ KD}} 
 & KDSNN  \cite{KDSNN}            & 67.18 \\
 & LaSNN   \cite{LaSNN}           & 66.94 \\
 & BKDSNN   \cite{BKDSNN}          & 67.21 \\
 & Logit-SNN   \cite{ensemble2}       & 71.04 \\
 & \cellcolor{pink!25}\textbf{SeAl-KD (Ours)} & \cellcolor{pink!25}\textbf{72.11} \\
\bottomrule
\end{tabular}
\caption{Comparison of different direct-training and distillation methods on ImageNet with ResNet-34.}
\label{tab:Imagenet}
\end{table}

\begin{table}[t]
\centering
\renewcommand{\arraystretch}{0.9}
\resizebox{\columnwidth}{!}{
\begin{tabular}{ccccc}
\toprule
 & \textbf{Method} & \textbf{Architecture} & \textbf{T} & \textbf{Acc(\%)} \\
\midrule
\multirow{6}{*}{\rotatebox[origin=c]{90}{w/o KD}}
 & STBP-tdBN \cite{STBP-tdBN} & ResNet-19 & 10 & 67.80  \\
\cmidrule(lr){2-5}
 & Dspike \cite{Dspike}    & ResNet-18 & 10 & 75.40  \\
\cmidrule(lr){2-5}
 & TET  \cite{TET}     & VGG-11    & 10 & 83.17 \\
\cmidrule(lr){2-5}
 & SLTT  \cite{SLTT}    & VGG-11    & 10 & 82.20  \\
\cmidrule(lr){2-5}
 & \multicolumn{2}{c}{Spikformer \cite{Spikformer} }               & 16 & 80.90 \\
\cmidrule(lr){2-5}
 & \multicolumn{2}{c}{Spike-driven Transformer \cite{Spike-driven} }  & 16 & 80.00   \\
\midrule
\multirow{9}{*}{\rotatebox[origin=c]{90}{w/ KD}} & \multirow{2}{*}{SAKD \cite{SAKD}} & VGG-11 & 4 & 81.50 \\
 &  & ResNet-19 & 4 & 80.30 \\
\cmidrule(lr){2-5}
 & \multirow{2}{*}{TSSD \cite{TSSD}} & \multirow{2}{*}{ResNet-18} & 8 & 72.90 \\
 &  &  & 16 & 81.60 \\
\cmidrule(lr){2-5}
 & \multirow{2}{*}{LogitSNN \cite{ensemble2}} & \multirow{2}{*}{ResNet-18} & 4 & 83.50 \\
 &  &  & 10 & 86.40 \\
\cmidrule(lr){2-5}
 & \cellcolor{pink!25} & \cellcolor{pink!25} & \cellcolor{pink!25} 4 & \cellcolor{pink!25} \textbf{84.00$_{\pm 0.20}$} \\
 & \cellcolor{pink!25} \multirow{-2}{*}{\textbf{SeAl-KD (Ours)}} & \cellcolor{pink!25} \multirow{-2}{*}{ResNet-18} & \cellcolor{pink!25} 10 & \cellcolor{pink!25} \textbf{86.70$_{\pm 0.20}$} \\
\bottomrule
\end{tabular}
}
\caption{Comparison of different direct-training and distillation methods on DVS-CIFAR10.}
\label{tab:DVS}
\end{table}

\begin{table}[t]
\centering
\setlength{\tabcolsep}{3pt}
\begin{tabular}{lccc}
\toprule
Method & CIFAR10 & CIFAR100 & DVS-C10 \\
\midrule
Timestep-wise KD      & 95.04  & 78.11 & 80.90 \\
w/ ELA    & 95.95 & 79.86 & \textbf{84.20} \\
w/ STA    & 95.60 & 78.76 & 83.80 \\
\rowcolor{pink!25}\textbf{SeAl-KD} & \textbf{96.00} & \textbf{80.01} & \textbf{84.20} \\
\bottomrule
\end{tabular}
\caption{Component ablation of SeAl-KD using ResNet-18 across datasets under $T=4$.}
\label{tab:com_ablation}
\end{table}

\begin{table}[t]
\centering
\small
\setlength{\tabcolsep}{6pt}
\begin{tabular}{lccc}
\toprule
\textbf{Method} & \textbf{CIFAR-10} & \textbf{CIFAR-100} & \textbf{DVS-CIFAR10} \\
\midrule
w/o ELA        & 95.04  & 78.11 & 80.9 \\
ELA-S          & 95.85 & 79.56 & \textbf{84.2} \\
ELA-A          & 95.78 & 79.34 & 83.6 \\
ELA-AS         & 95.76 & 79.29 & 82.9 \\
ELA-Both       & 95.82 & 79.36 & 83.9 \\
\rowcolor{pink!25} \textbf{ELA (Ours)} & \textbf{95.95} & \textbf{79.86} & \textbf{84.2} \\
\bottomrule
\end{tabular}
\caption{Ablation study of ELA variants using ResNet-18 across different datasets under $T=4$.}
\label{tab:ela_ablation}
\end{table}

\begin{table}[t]
\centering
\small
\setlength{\tabcolsep}{6pt}
\begin{tabular}{lccc}
\toprule
\textbf{Method} & \textbf{CIFAR-10} & \textbf{CIFAR-100} & \textbf{DVS-CIFAR10} \\
\midrule
w/o STA        & 95.04  & 78.11 & 80.9 \\
UTA          & 95.55 & 78.58 & 82.4 \\
STA-NoConf     & 95.32 & 78.57 & \textbf{84.1} \\
STA-NoSim      & 95.58 & 78.60  & 83.6 \\
STA-Dist       & 95.37 & 78.60  & 83.7 \\
\rowcolor{pink!25}\textbf{STA (Ours)} 
& \textbf{95.60} 
& \textbf{78.76} 
& 83.8 \\
\bottomrule
\end{tabular}
\caption{Ablation study of STA variants using ResNet-18 across different datasets under $T=4$.}
\label{tab:sta_ablation}
\end{table}

\subsection{Ablation Study}
\subsubsection{Component Ablation of SeAl-KD}
We first evaluate the individual and joint contributions of ELA and STA in Table~\ref{tab:com_ablation}. Removing either module degrades performance, and removing both causes a larger drop. This indicates that SeAl-KD benefits from making alignment selective in both class and temporal dimensions: ELA reduces misleading supervision on erroneous classes, while STA provides temporal guidance from more reliable states.

\subsubsection{Ablation Study of ELA Variants}
We analyze several ELA variants, with results reported in Table~\ref{tab:ela_ablation}. \textbf{ELA-S} applies error-aware modification only to the student logits, \textbf{ELA-A} only to the teacher logits, \textbf{ELA-AS} introduces error awareness on the ANN teacher side and aligns the student accordingly, and \textbf{ELA-Both} extends the alignment to three classes when the teacher prediction is also incorrect. All variants outperform the baseline without ELA, showing that class-level selective correction is beneficial once alignment is no longer imposed uniformly. Our ELA achieves the best overall performance, and student-driven variants are consistently more effective than teacher-driven ones, because student errors directly identify the timesteps where naive alignment is most misleading. This supports that ELA should be selective, student-driven, and focused on correcting the dominant confusing class relation rather than enforcing broader per-timestep matching.

\setcounter{figure}{4}
\begin{figure*}[t]
    \centering
    \includegraphics[width=0.95\textwidth]{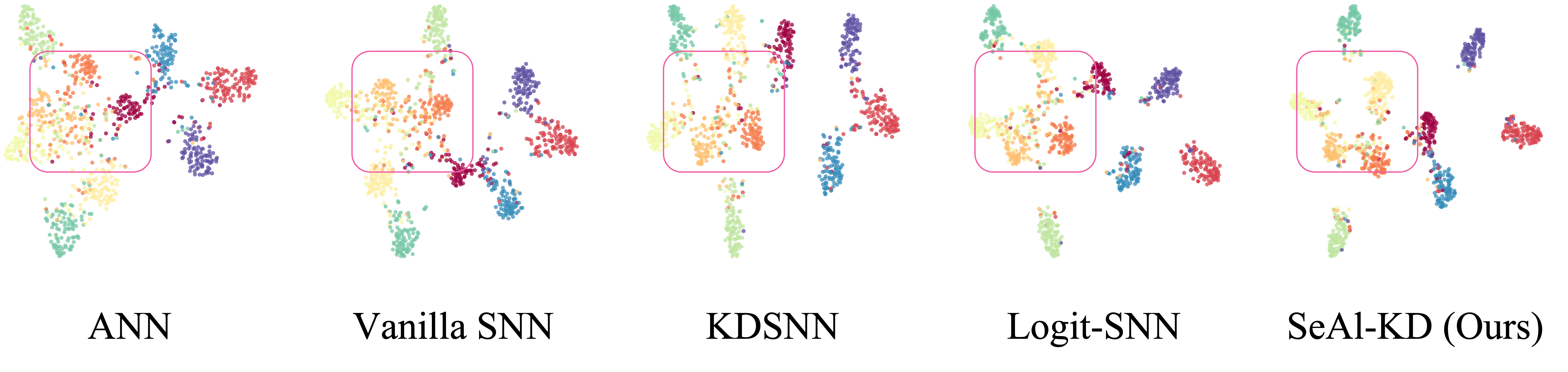}
    \caption{t-SNE visualization of learned feature representations on DVS-CIFAR10 under different direct-training and distillation methods.}
    \label{fig:sne}
\end{figure*}

\subsubsection{Ablation Study of STA Variants}
We analyze several STA variants, with results reported in Table~\ref{tab:sta_ablation}. \textbf{STA-NoConf} removes confidence weighting, \textbf{STA-NoSim} removes similarity filtering, \textbf{STA-Dist} replaces cosine similarity with cosine distance, and \textbf{UTA} enforces unweighted pairwise alignment across all timesteps. All variants outperform the baseline without STA, confirming the benefit of temporal guidance, while the full STA performs best overall. This suggests that temporal correction should not be propagated uniformly across timesteps, but should instead guide erroneous timesteps using temporal states that are both reliable and compatible. Confidence weighting suppresses noisy sources, similarity filtering avoids mismatched guidance, and their combination yields the most effective selective temporal correction.

\subsection{Visualization}
\subsubsection{Temporal Discrepancy}
\begin{figure}[t]
    \centering
    \includegraphics[width=1\linewidth]{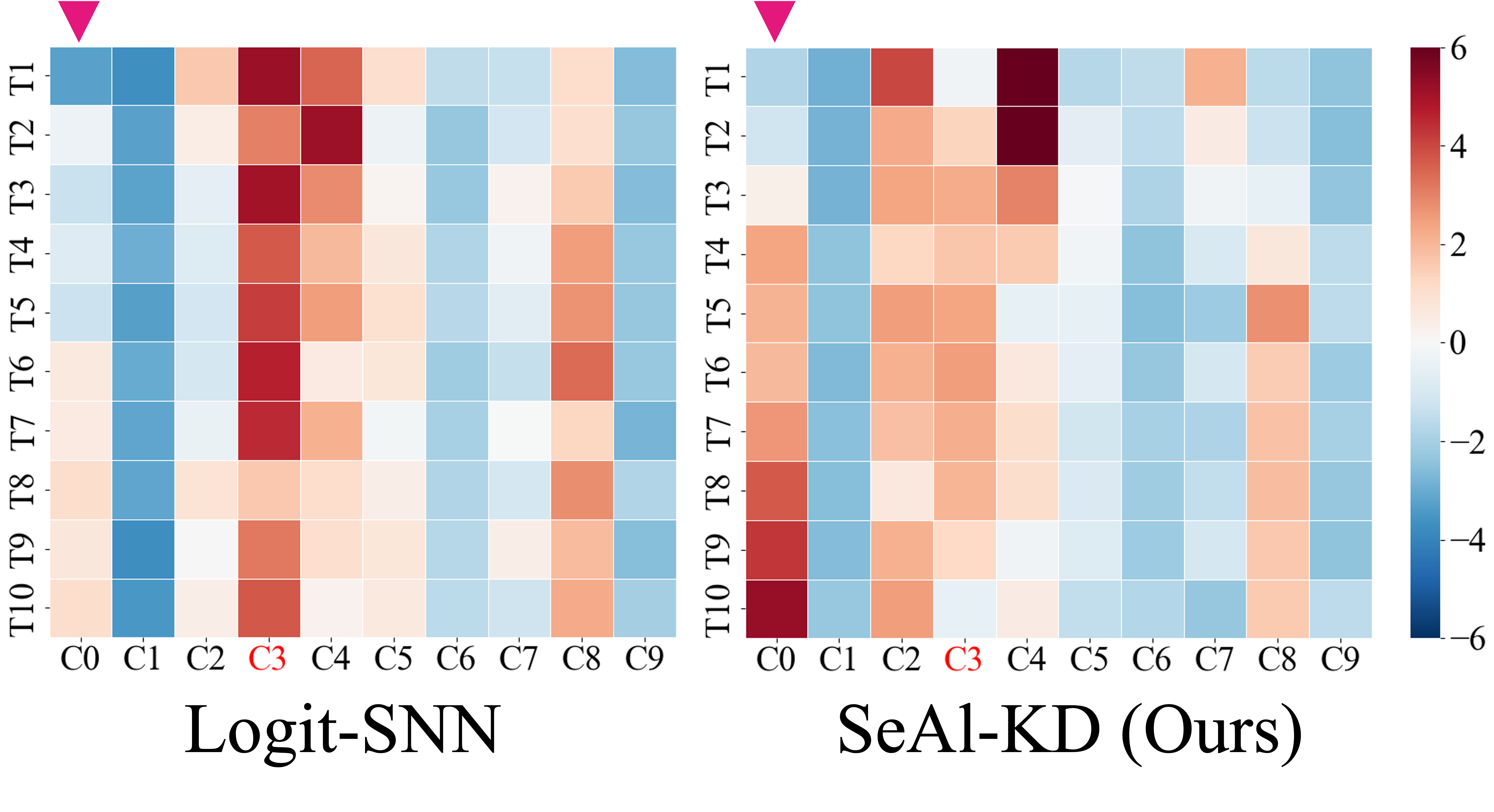}
    \caption{Heatmap of per-timestep class logits on DVS-CIFAR10 for Logit-SNN and SeAI-KD.}
    \label{fig:heatmap}
\end{figure}

To better understand how selective alignment affects temporal prediction dynamics, we visualize per-timestep class logits on DVS-CIFAR10 in Figure~\ref{fig:heatmap}. Compared with Logit-SNN, which exhibits a largely time-invariant pattern where a subset of classes remains dominant across most timesteps, our method produces more structured temporal trajectories. The dominant class can shift over time, and certain classes show larger logit variation across timesteps. For example, given that the ground-truth label is $C$3, the logit of $C$0 transitions from negative to strongly positive. Overall, the visualization suggests that SeAI-KD redistributes evidence over time and encourages temporally differentiated class responses, rather than maintaining a fixed class preference across timesteps.

\subsubsection{t-SNE}
To better understand the effect of selective temporal alignment, we visualize the learned representations using t-SNE, as shown in Figure~\ref{fig:sne}. Compared with direct training and previous distillation methods, our approach produces more compact class clusters with clearer separation. Notably, the circled regions exhibit less class overlap and fewer scattered samples. Overall, the visualization indicates that selective alignment leads to more discriminative and well-structured representations over time.

\section{Conclusion}
This work examines the limitations of uniform timestep-wise KD in SNNs and shows that treating all temporal states equally can impose overly rigid supervision, even though intermediate timesteps need not all be individually correct. Instead, erroneous timesteps should receive guidance that moves them toward the correct final outcome. We propose SeAl-KD, which selectively aligns class-level and temporal knowledge via ELA and STA. Our analysis shows that ELA and STA reduce the influence of erroneous, low-confidence, or incompatible timesteps during alignment, enabling more effective correction where it is needed. Experiments on both static and neuromorphic datasets further confirm that this selective alignment strategy consistently improves accuracy with only minimal computational overhead.

\clearpage
\section*{Ethical Statement}
There are no ethical issues.

\section*{Acknowledgements}
This work was supported by start-up funds with No. MSRI8001004 and No. MSRI9002005 and Monash eResearch capabilities, including M3.

\bibliographystyle{named}
\bibliography{ijcai26}

\clearpage
\appendix

\begin{strip}
    \centering
    \includegraphics[width=\textwidth]{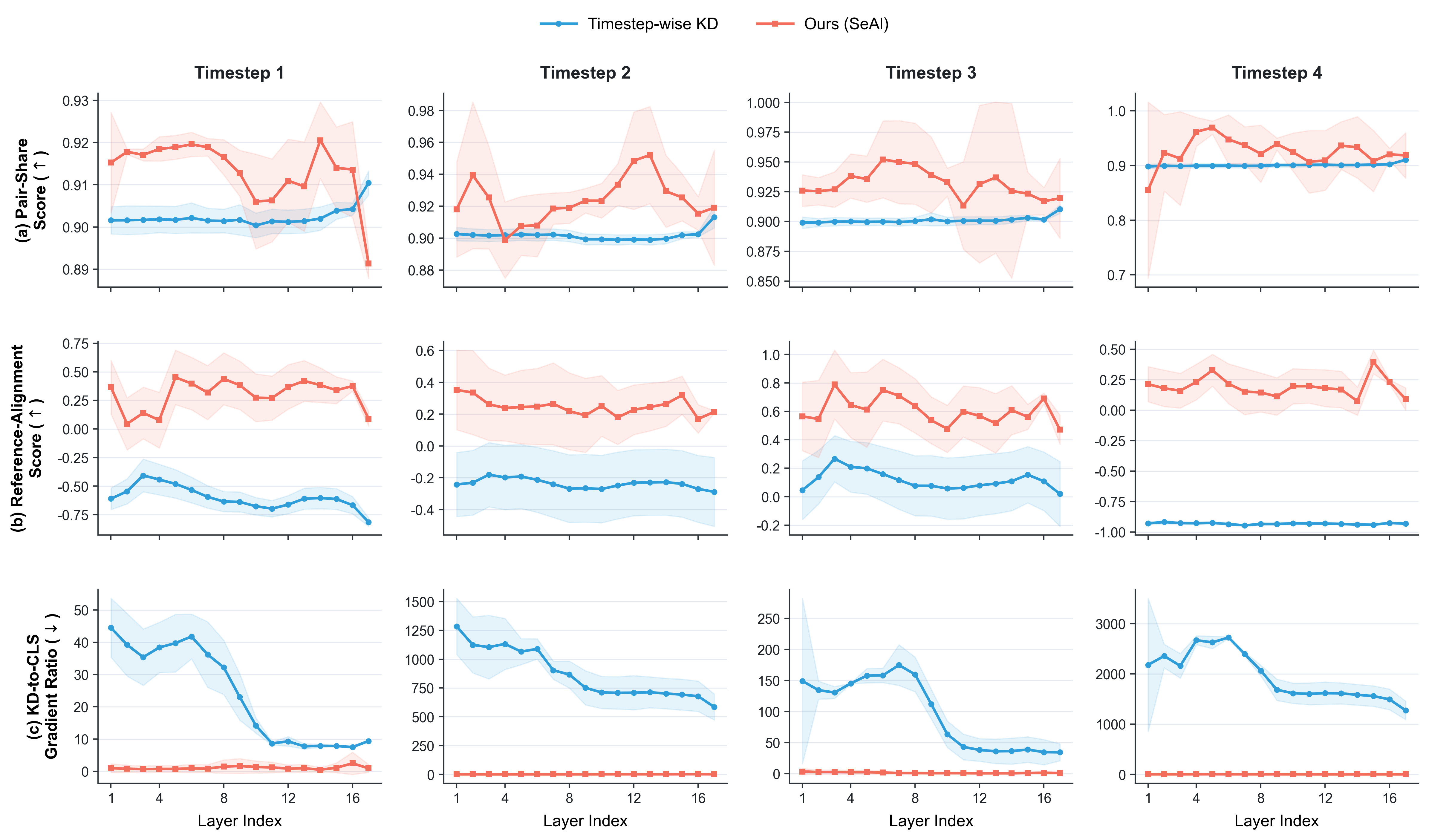}
    \captionof{figure}{Layer-wise statistics over all timesteps for the three propositions: (a) the fraction of the ELA update assigned to the ground-truth class and the dominant false class at erroneous timesteps; (b) the cosine similarity between the STA update and the direction that reduces the gap to reliability-weighted temporal references at weak timesteps; (c) the ratio between the distillation-gradient norm and the classification-gradient norm at already-correct timesteps. Statistics are computed from five randomly selected samples and reported as mean $\pm$ std.}
    \label{fig:analysis_appendix}
\end{strip}

\section{Structural Temporal Fluctuations in LIF Dynamics}
\label{app:lif_variance}
The leak--integrate--reset mechanism of LIF neurons inherently produces temporal
fluctuations in membrane potentials, even under constant input. Consider a single
LIF neuron receiving a fixed current $I$. Its membrane potential evolves as
\begin{equation}
u_{t+1} = \alpha u_t + I - V_{\mathrm{th}}\, s_t, 
\qquad s_t \in \{0,1\},
\end{equation}
where $0 < \alpha < 1$ is the leak factor and $V_{\mathrm{th}} > 0$ is the firing threshold.
In any non-trivial firing regime (neither always silent nor always firing), the
neuron exhibits both spike and non-spike timesteps. Consequently, the update increments
take two distinct values:
\begin{equation}
u_{t+1} - u_t = (\alpha - 1)\,u_t + I \qquad \text{(no spike)},
\end{equation}
\begin{equation}
u_{t+1} - u_t = (\alpha - 1)\,u_t + I - V_{\mathrm{th}} 
\qquad \text{(spike)}.
\end{equation}

These two increments differ by exactly $V_{\mathrm{th}} > 0$, so the sequence
$\{u_t\}$ cannot remain constant over time and must visit at least two distinct
membrane-potential levels with non-zero frequency. This implies a strictly positive
temporal variance, showing that temporal fluctuations arise naturally from the LIF
update rule.

\begin{table*}[t]
\centering
\begin{tabular}{cccccccc}
\toprule
\multirow{2}{*}{\textbf{Dataset}} &
\multirow{2}{*}{\textbf{\shortstack{Batch\\Size}}} &
\multirow{2}{*}{\textbf{Epochs}} &
\multirow{2}{*}{\textbf{\shortstack{Learning\\Rate}}} &
\multirow{2}{*}{\textbf{\shortstack{Weight\\Decay}}} &
\multirow{2}{*}{\textbf{\shortstack{Student\\Architecture}}} &
\multirow{2}{*}{\textbf{\shortstack{Teacher\\Architecture}}} &
\multirow{2}{*}{\textbf{\shortstack{Teacher\\Acc (\%)}}} \\
 & & & & & & & \\
\midrule
\multirow{2}{*}{CIFAR-10}
 & \multirow{2}{*}{128} & \multirow{2}{*}{300} & \multirow{2}{*}{0.1} & \multirow{2}{*}{5e-4}
 & ResNet-18 & ResNet-34 & 97.06 \\
 &  &  &  &  & ResNet-19 & ResNet-19 & 97.20 \\
\midrule
\multirow{2}{*}{CIFAR-100}
 & \multirow{2}{*}{128} & \multirow{2}{*}{300} & \multirow{2}{*}{0.1} & \multirow{2}{*}{5e-4}
 & ResNet-18 & ResNet-34 & 81.31 \\
 &  &  &  &  & ResNet-19 & ResNet-19 & 82.57 \\
\midrule
\multirow{2}{*}{DVS-CIFAR10}
 & \multirow{2}{*}{32}
 & \multirow{2}{*}{300}
 & \multirow{2}{*}{0.1}
 & \multirow{2}{*}{5e-4}
 & \multirow{2}{*}{ResNet-18}
 & ResNet-19-T4
 & 83.80 \\
 &  &  &  &  &  & ResNet-19-T10 & 83.60 \\
\midrule
ImageNet
 & 512 & 100 & 0.2 & 2e-5
 & ResNet-34 & ResNet-34 & 71.24 \\
\bottomrule
\end{tabular}
\caption{
Training settings across different datasets.
}
\label{tab:training_datails}
\end{table*}

\section{Theoretical Analysis Across Timesteps}
\label{app:theoretical_figures}
This section extends the theoretical and statistical analysis in the main text by reporting the corresponding layer-wise statistics over all timesteps. Together with Figure~\ref{fig:analysis}, these results show that the trends supporting the three propositions remain consistent across timesteps.

\section{Dataset Details}
\label{app:dataset-details}
\subsection{Static Image Datasets}
\textbf{CIFAR-10 and CIFAR-100.}
CIFAR-10 and CIFAR-100 \cite{cifar} are image classification datasets consisting of 60{,}000
color images with a spatial resolution of $32 \times 32$. Each dataset contains
50{,}000 training images and 10{,}000 test images. CIFAR-10 includes 10 object
classes, while CIFAR-100 contains 100 classes with finer granularity. Both
datasets share the same image format and data split. Standard preprocessing and
data augmentation are applied during training.

\noindent\textbf{ImageNet.}
ImageNet \cite{imagenet} is an image classification dataset containing approximately 1.28
million training images and 50{,}000 validation images spanning 1{,}000 object
categories. The images have varying spatial resolutions and are resized and
cropped during preprocessing to match the input size required by the models.

\subsection{Neuromorphic Dataset}

\textbf{DVS-CIFAR10.}
DVS-CIFAR10 \cite{dvscifar} is a neuromorphic dataset derived from CIFAR-10 and recorded using a
Dynamic Vision Sensor. Instead of frame-based images, the dataset represents
visual information as asynchronous streams of events, where each event is
described by its spatial location, timestamp, and polarity of brightness
changes. It contains 10 object classes corresponding to CIFAR-10. For
training and evaluation, the event streams are converted into discrete-time
representations by accumulating events within fixed temporal windows.

\section{Training Details}
\label{app:training-details}
All experiments are optimized using stochastic gradient descent with momentum set to 0.9. The learning rate is scheduled by a cosine decay strategy throughout training. All implementations are based on the PyTorch framework.
For static image benchmarks, including CIFAR and ImageNet, different training configurations are adopted according to dataset scale. Models on CIFAR datasets are trained using a single NVIDIA A100 GPU. For ImageNet, distributed data parallelism is employed across eight A100 GPUs. 
For event-based vision tasks, experiments on the DVS-CIFAR10 dataset follow the same optimization strategy and are conducted on a single A100 GPU. Input event streams are processed into frame-based representations before being fed into the network. Detailed network configurations and all hyperparameter choices for different datasets and model variants are provided in Table~\ref{tab:training_datails}.

\begin{figure*}[t]
    \centering
    \begin{subfigure}[t]{0.32\textwidth}
        \centering
        \includegraphics[width=\linewidth]{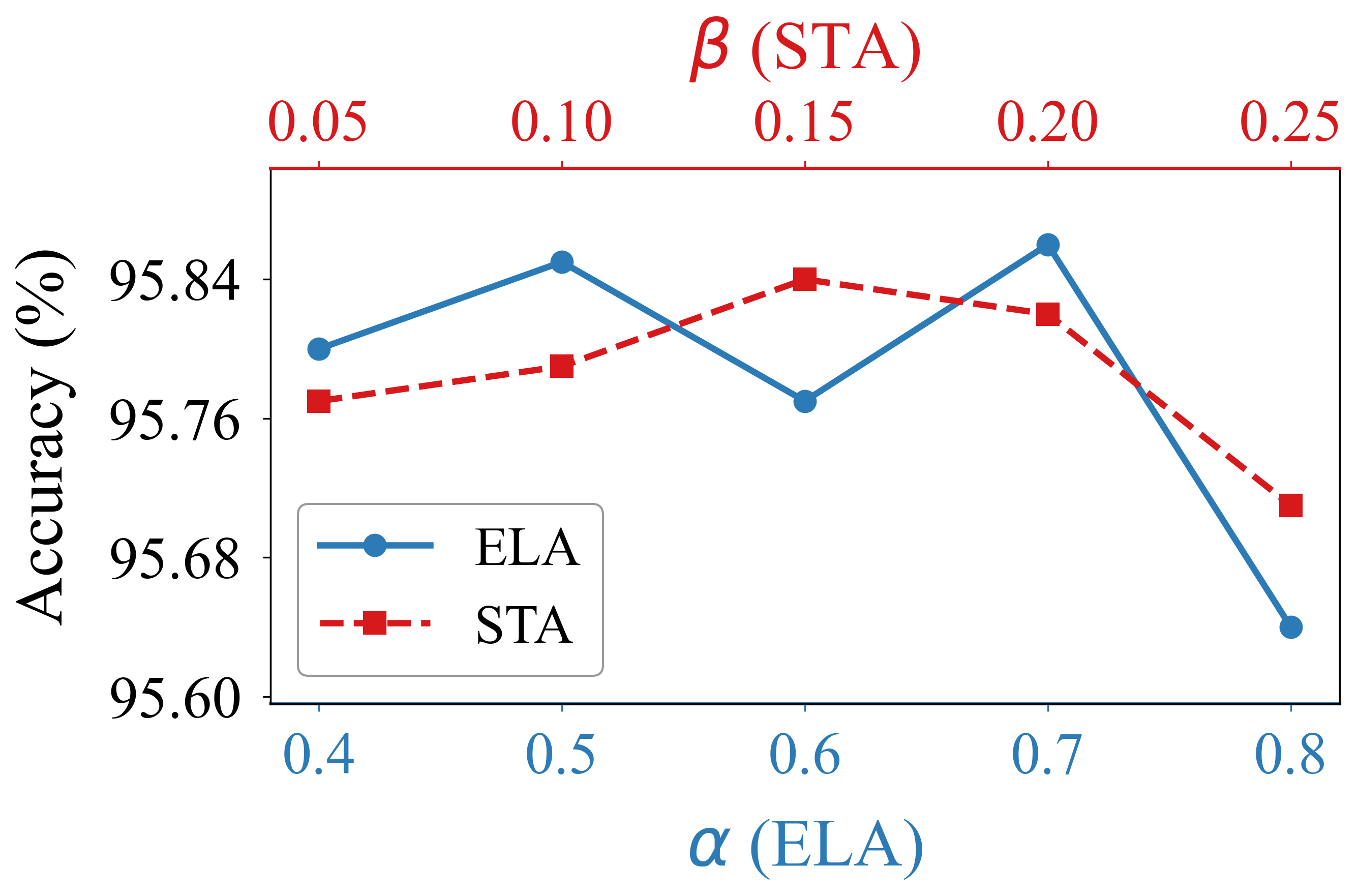}
        \caption{CIFAR-10.}
    \end{subfigure}
    \hfill
    \begin{subfigure}[t]{0.32\textwidth}
        \centering
        \includegraphics[width=\linewidth]{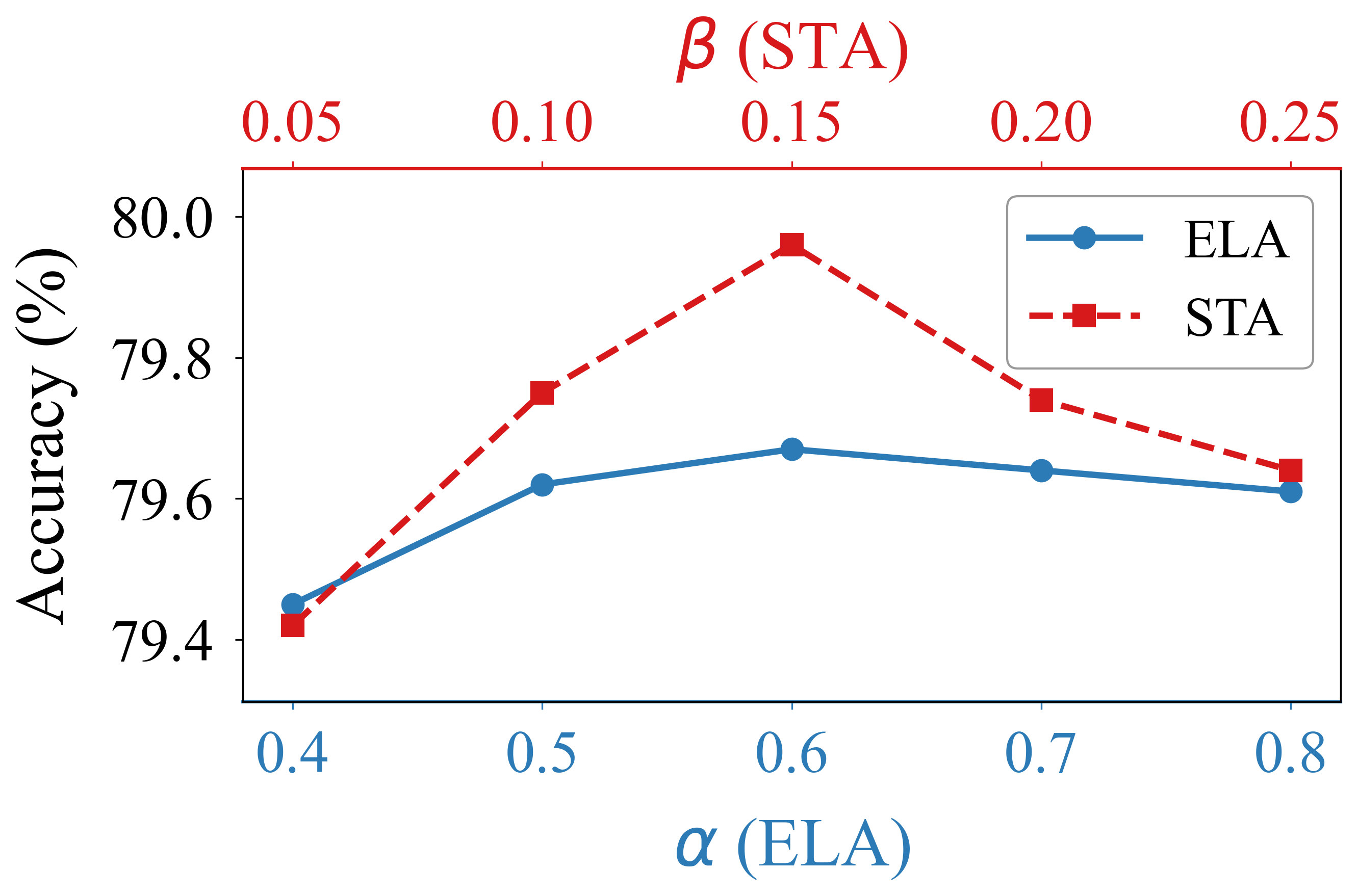}
        \caption{CIFAR-100.}
    \end{subfigure}
    \hfill
    \begin{subfigure}[t]{0.32\textwidth}
        \centering
        \includegraphics[width=\linewidth]{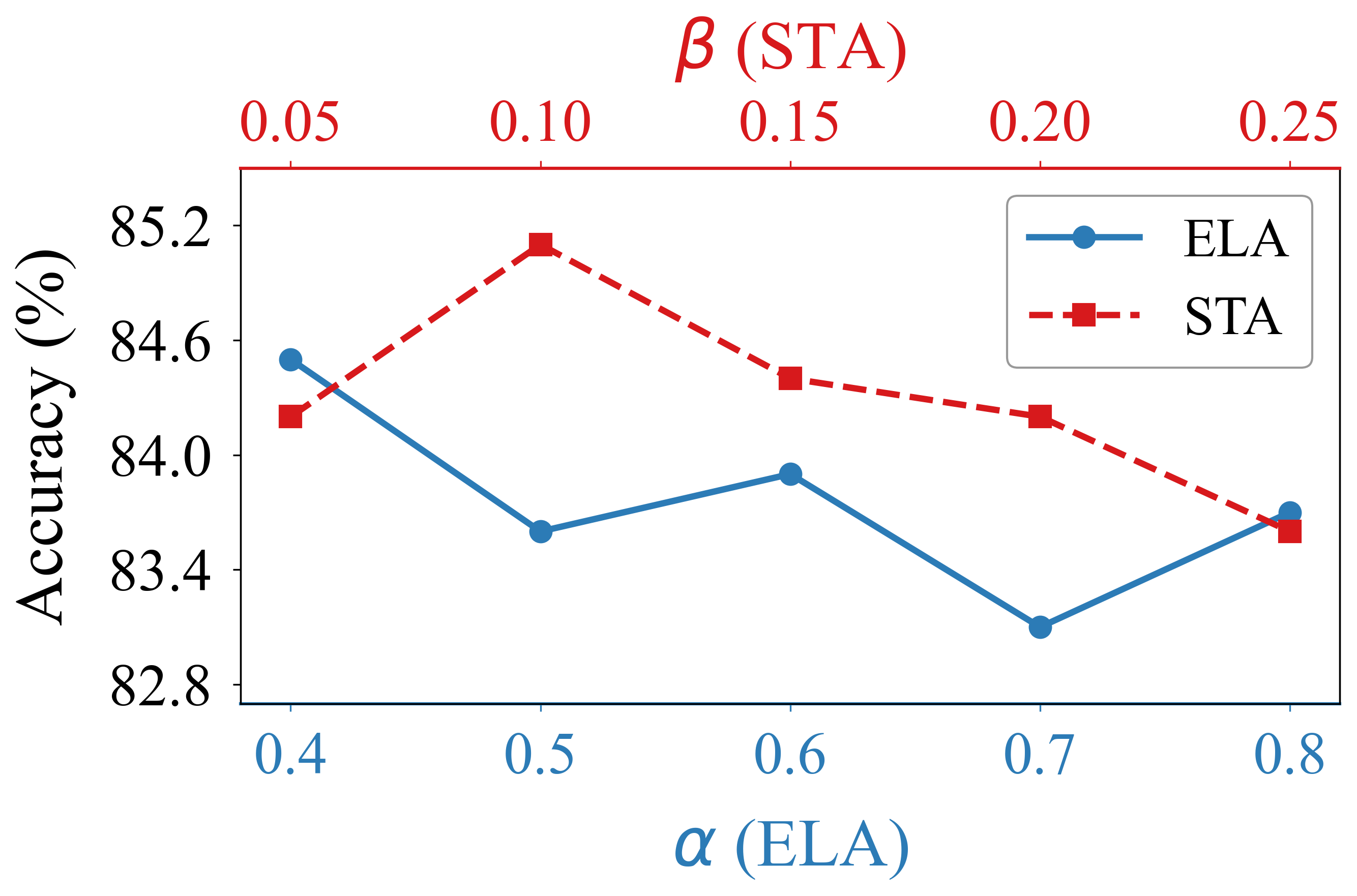}
        \caption{DVS-CIFAR10.}
    \end{subfigure}
    \caption{Sensitivity analysis of hyperparameters on three datasets under $T=4$.}
    \label{fig:sensitivity_all}
\end{figure*}

\section{Energy Consumption Analysis}
\label{app:energy}
To quantify the computational energy cost of SNNs, we follow a commonly adopted evaluation protocol in neuromorphic computing, which characterizes energy consumption in terms of synaptic operations~\cite{Spikformer}. Specifically, the overall synaptic operation power (SOP) is modeled as the weighted sum of accumulation and multiply-accumulate operations:
\begin{equation}
\text{SOP}_s = E_{\mathrm{AC}} \cdot AC_s + E_{\mathrm{MAC}} \cdot MAC_s,
\end{equation}
where $AC_s$ and $MAC_s$ denote the total numbers of accumulation (AC) and multiply-accumulate (MAC) operations, respectively. The coefficients $E_{\mathrm{AC}}$ and $E_{\mathrm{MAC}}$ correspond to the energy cost of a single operation of each type. Following the hardware energy model introduced in~\cite{han2015learning}, a 32-bit floating-point addition is assumed to consume $0.9$ picojoules (pJ), while a 32-bit MAC operation requires $4.6$ pJ.

In SNNs, information is conveyed through discrete spike events. Let $s_i^l[t] \in \{0,1\}$ indicate whether neuron $i$ in layer $l$ emits a spike at timestep $t$. Whenever a spike is generated, all outgoing synapses of that neuron are activated, and each activated synapse contributes one accumulation operation. If neuron $i$ in layer $l$ has $f_i^l$ outgoing connections, the total number of AC operations accumulated over the entire network and all timesteps can be written as:
\begin{equation}
AC_s = \sum_{t=1}^{T} \sum_{l=1}^{L-1} \sum_{i=1}^{N^l} f_i^l \, s_i^l[t],
\end{equation}
where $T$ denotes the simulation length in timesteps, $L$ is the total number of layers, and $N^l$ is the number of neurons in the $l$-th layer.

By contrast, artificial neural networks (ANNs) do not exhibit temporal spiking behavior. Each neuron performs a single feedforward computation, and every synaptic connection contributes exactly one MAC operation. Therefore, the total number of MAC operations in an ANN is solely determined by the network connectivity:
\begin{equation}
MAC_s = \sum_{l=1}^{L-1} \sum_{i=1}^{N^l} f_i^l.
\end{equation}

Using the above formulations, the SOP of both SNNs and ANNs can be consistently estimated by combining the corresponding operation counts with their associated per-operation energy costs.

Table~\ref{tab:energy} indicates that energy is primarily driven by ACs, while MACs remain essentially constant under the same architecture. For each dataset, we report results using the maximum number of timesteps $T$ adopted in this paper. Compared with ANNs, SNNs can therefore achieve substantially lower energy consumption because most computations shift away from MAC-dominated processing and become event-driven. Under this setting, the energy gap across different SNN methods is largely explained by how many spikes they produce, since firing activity directly determines the number of ACs. Our method does not increase the fire rate or energy consumption. It maintains comparable energy and can even reduce it for certain architecture–dataset combinations by lowering the fire rate relative to LogitSNN, thereby reducing ACs without sacrificing accuracy. In addition, the training-time overhead is negligible, suggesting that the method introduces only lightweight operations during optimization.

\section{Hyperparameters Sensitivity Analysis}
\label{app:sensitivity}

We analyze the sensitivity of the proposed method to the weighting coefficients $\alpha$ for ELA and $\beta$ for STA. 
We first set $\beta = 0$ and vary $\alpha$ to evaluate ELA alone, and then fix $\alpha = 0.6$ and vary $\beta$ to evaluate STA on top of ELA. 
Figure~\ref{fig:sensitivity_all} reports the results on CIFAR-10, CIFAR-100, and DVS-CIFAR10. 

The performance changes are small across a reasonable range of coefficients, and the improvements over the corresponding baseline remain consistent. 
These results indicate that the proposed method is relatively insensitive to the choice of $\alpha$ and $\beta$ and does not require careful tuning. 
Based on this analysis, we use $\alpha = 0.6$ and $\beta = 0.15$ in all experiments.

\begin{table*}[t]
\centering
\begin{tabular}{ccccccccc}
\toprule
\multirow{2}{*}{\textbf{Data}} & \multirow{2}{*}{\textbf{Architecture}} & \multirow{2}{*}{\textbf{Method}} & \textbf{GPU h} & \textbf{Fire Rate} & \textbf{ACs} & \textbf{MACs} & \textbf{Energy} & \textbf{Acc} \\
 &  &  & \textbf{(h)} & \textbf{(\%)} & \textbf{(M)} & \textbf{(M)} & \textbf{($\mu$J)} & \textbf{(\%)} \\ 
\midrule
\multirow{5}{*}[-2ex]{CIFAR-10} & \multirow{5}{*}[-2ex]{Resnet18} & ANN & 0.52 & - & 0.56 & 549.13 & 2526.50 & 97.06 \\ \cmidrule{3-9}
 &  & VanillaSNN & 4.66 & 14.33 & 73.12 & 3.34 & 80.62 & 95.66 \\ \cmidrule{3-9}
 &  & KDSNN & 4.90 & 14.83 & 75.85 & 3.34 & 86.15 & 95.78 \\ \cmidrule{3-9}
 &  & LogitSNN & 5.06 & 15.74 & 82.26 & 3.34 & 90.90 & 96.12 \\ \cmidrule{3-9}
 &  & \cellcolor{pink!25}SeAl (Ours) & \cellcolor{pink!25}5.07 & \cellcolor{pink!25}14.89 & \cellcolor{pink!25}81.56 & \cellcolor{pink!25}3.34 & \cellcolor{pink!25}88.78 & \cellcolor{pink!25}96.18 \\ \midrule

\multirow{5}{*}[-2ex]{CIFAR-100} & \multirow{5}{*}[-2ex]{Resnet18} & ANN & 0.52 & - & 0.56 & 549.18 & 2526.72 & 81.31 \\ \cmidrule{3-9}
 &  & VanillaSNN & 4.64 & 17.45 & 92.92 & 3.34 & 99.82 & 78.33 \\ \cmidrule{3-9}
 &  & KDSNN & 4.82 & 17.99 & 93.65 & 3.34 & 101.93 & 79.31 \\ \cmidrule{3-9}
 &  & LogitSNN & 5.08 & 18.55 & 99.49 & 3.34 & 108.27 & 80.07 \\ \cmidrule{3-9}
 &  & \cellcolor{pink!25}SeAl (Ours) & \cellcolor{pink!25}5.08 & \cellcolor{pink!25}17.48 & \cellcolor{pink!25}98.33 & \cellcolor{pink!25}3.34 & \cellcolor{pink!25}103.87 & \cellcolor{pink!25}80.25 \\ \midrule

\multirow{5}{*}[-2ex]{CIFAR-10} & \multirow{5}{*}[-2ex]{Resnet19} & ANN & 1.17 & - & 1.44 & 2268.60 & 10436.84 & 97.20 \\ \cmidrule{3-9}
 &  & VanillaSNN & 11.66 & 13.11 & 286.39 & 8.65 & 296.99 & 96.73 \\ \cmidrule{3-9}
 &  & KDSNN & 11.92 & 12.24 & 271.88 & 8.65 & 285.06 & 96.98 \\ \cmidrule{3-9}
 &  & LogitSNN & 12.12 & 15.42 & 317.46 & 8.65 & 319.52 & 97.04 \\ \cmidrule{3-9}
 &  & \cellcolor{pink!25}SeAl (Ours) & \cellcolor{pink!25}12.14 & \cellcolor{pink!25}15.14 & \cellcolor{pink!25}310.54 & \cellcolor{pink!25}8.65 & \cellcolor{pink!25}319.77 & \cellcolor{pink!25}97.23 \\ \midrule

\multirow{5}{*}[-2ex]{CIFAR-100} & \multirow{5}{*}[-2ex]{Resnet19} & ANN & 1.17 & - & 1.44 & 2268.62 & 10436.95 & 82.57 \\ \cmidrule{3-9}
 &  & VanillaSNN & 11.79 & 16.15 & 350.70 & 8.65 & 355.43 & 81.11 \\ \cmidrule{3-9}
 &  & KDSNN & 12.08 & 16.26 & 359.65 & 8.65 & 363.49 & 82.12 \\ \cmidrule{3-9}
 &  & LogitSNN & 12.45 & 17.41 & 371.74 & 8.65 & 372.57 & 83.12 \\ \cmidrule{3-9}
 &  & \cellcolor{pink!25}SeAl (Ours) & \cellcolor{pink!25}12.46 & \cellcolor{pink!25}17.05 & \cellcolor{pink!25}366.02 & \cellcolor{pink!25}8.65 & \cellcolor{pink!25}369.22 & \cellcolor{pink!25}83.37 \\ \midrule

\multirow{5}{*}[-2ex]{DVS-Cifar10} & \multirow{5}{*}[-2ex]{Resnet18} & ANN & 1.04 & - & 0.56 & 549.13 & 2526.50 & 83 \\ \cmidrule{3-9}
 &  & VanillaSNN & 4.16 & 12.68 & 807.9 & 5.57 & 752.73 & 84 \\ \cmidrule{3-9}
 &  & KDSNN & 4.28 & 17.21 & 1090.37 & 5.57 & 1006.96 & 84 \\ \cmidrule{3-9}
 &  & LogitSNN & 4.41 & 15.65 & 1030.17 & 5.57 & 952.78 & 85.6 \\ \cmidrule{3-9}
 &  & \cellcolor{pink!25}SeAl (Ours) & \cellcolor{pink!25}4.45 & \cellcolor{pink!25}14.84 & \cellcolor{pink!25}969.08 & \cellcolor{pink!25}5.57 & \cellcolor{pink!25}897.8 & \cellcolor{pink!25}86.7 \\ 
\bottomrule
\end{tabular}
\caption{Comparison of energy consumption, accumulation (ACs) and multiply–accumulate (MACs) operations, training cost (GPU hours), and accuracy on CIFAR-10 and CIFAR-100 with ResNet-18/ResNet-19 at $T=6$, and on DVS-CIFAR10 with ResNet-18 at $T=10$.}
\label{tab:energy}
\end{table*}

\section{Use of Large Language Models}
Large language models were used solely for language editing (e.g., grammar and clarity). All ideas, experiments, and analyses are entirely the authors’ own.

\end{document}